\documentclass[lettersize,journal]{IEEEtran}
\usepackage{amsmath,amsfonts}
\usepackage{array}
\usepackage{textcomp}
\usepackage{stfloats}
\usepackage{url}
\usepackage{tikz}
\usepackage{verbatim}
\usepackage{graphicx}
\def\BibTeX{{\rm B\kern-.05em{\sc i\kern-.025em b}\kern-.08em
    T\kern-.1667em\lower.7ex\hbox{E}\kern-.125emX}}
\usepackage{balance}

\usepackage{graphics} 
\usepackage{epsfig} 
\usepackage{mathptmx} 
\usepackage{times} 
\usepackage{amsmath} 
\usepackage{amssymb}  
\usepackage{siunitx}
\usepackage{hyperref}
\usepackage{multirow}
\usepackage{multicol}
\usepackage{nicefrac}
\usepackage{svg}
\usepackage{graphicx}
\usepackage{cite}
\usepackage{picinpar}
\usepackage[latin1]{inputenc}
\usepackage{colortbl}
\usepackage{soul}
\usepackage{multirow}
\usepackage{pifont}
\usepackage{color}
\usepackage{alltt}
\usepackage{enumerate}
\usepackage{siunitx}
\usepackage{epstopdf}
\usepackage{pbox}
\usepackage{amsfonts}
\usepackage{textcomp}
\usepackage{diagbox}
\usepackage{graphics} 
\usepackage{times} 
\usepackage{amssymb}  
\usepackage{microtype}
\usepackage{subfigure}
\usepackage{booktabs} 
\usepackage{xfrac}
\usepackage{nicefrac}
\usepackage{multicol}
\usepackage{array}
\usepackage{algorithm}
\usepackage{gensymb}
\usepackage{xr}
\usepackage{algpseudocode}

\usepackage{makecell}

\usepackage{ulem}

\DeclareMathOperator*{\argmin}{arg\,min}

\def\BibTeX{{\rm B\kern-.05em{\sc i\kern-.025em b}\kern-.08em
    T\kern-.1667em\lower.7ex\hbox{E}\kern-.125emX}}

\begin{document}

\title{Deep Reinforcement Learning-Based DRAM Equalizer Parameter Optimization Using Latent Representations}

\author{Muhammad Usama and
	Dong Eui Chang\\
	Control Laboratory, School of Electrical Engineering, KAIST, \\
	Daejeon, 34141, Republic of Korea
	\thanks{M. Usama and D. E. Chang are with the Control Laboratory, School of Electrical Engineering, KAIST, Daejeon, 34141, Republic of Korea (e-mail: usama@kaist.ac.kr; dechang@kaist.ac.kr).}
}




\maketitle

\begin{abstract}
Equalizer parameter optimization for signal integrity in high-speed Dynamic Random Access Memory systems is crucial but often computationally demanding or model-reliant. This paper introduces a data-driven framework employing learned latent signal representations for efficient signal integrity evaluation, coupled with a model-free Advantage Actor-Critic reinforcement learning agent for parameter optimization. The latent representation captures vital signal integrity features, offering a fast alternative to direct eye diagram analysis during optimization, while the reinforcement learning agent derives optimal equalizer settings without explicit system models. Applied to industry-standard Dynamic Random Access Memory waveforms, the method achieved significant eye-opening window area improvements: 42.7\% for cascaded Continuous-Time Linear Equalizer and Decision Feedback Equalizer structures, and 36.8\% for Decision Feedback Equalizer-only configurations. These results demonstrate superior performance, computational efficiency, and robust generalization across diverse Dynamic Random Access Memory units compared to existing techniques. Core contributions include an efficient latent signal integrity metric for optimization, a robust model-free reinforcement learning strategy, and validated superior performance for complex equalizer architectures.
\end{abstract}


\section{Introduction}
\IEEEPARstart{T}{he} increasing demand for higher data rates in modern Dynamic Random Access Memory (DRAM) systems presents significant challenges in maintaining robust signal integrity (SI). As data rates rise, signal distortions, most notably inter-symbol interference (ISI), become increasingly severe, degrading the quality of signals received by DRAM modules. Equalizers, such as continuous-time linear equalizers (CTLE), feed-forward equalizers, and decision feedback equalizers (DFE), are essential for compensating channel impairments and restoring SI. However, the optimization of equalizer parameters is a complex and critical task that directly impacts system reliability and performance.

Traditional equalizer optimization methods rely on eye diagram analysis combined with exhaustive or heuristic parameter search algorithms, such as genetic algorithms, to determine optimal settings for mitigating ISI and channel loss. While effective, these approaches are computationally intensive and lack adaptability to dynamic channel conditions. Adaptive algorithms based on the least-mean-square criterion~\cite{LMS_REF1, LMS_REF2} offer computational efficiency but require accurate channel models and explicit error signals, and may converge to suboptimal local minima in channels with severe ISI~\cite{LMS_REF3}. Alternative heuristic~\cite{heuristics1, heuristics2} and iterative~\cite{iterative1, iterative2} optimization strategies have also been explored, but their ability to generalize across diverse operational scenarios is limited~\cite{cooptimized1, cooptimized2}.

Recent advances in machine learning have introduced new paradigms for equalizer parameter optimization. Surrogate modeling, using deep neural networks and support vector regression, has accelerated SI analysis and parameter estimation~\cite{wu2024dnn, shi2021surrogate, hui2024cnn}. Autoencoders and generative models have been used to compress eye diagram information into latent codes, enabling efficient SI metrics~\cite{Song2019LearningPC, Song2021ModelBasedEL}. Reinforcement learning (RL), particularly actor-critic methods, have been applied to equalizer optimization in high-bandwidth memory links, demonstrating improvements over traditional search-based methods~\cite{choi2021rlpeq, choi2022hybrid, kaist-reference, kaist-ddpg}. The integration of latent-space representations with RL frameworks has further enhanced learning stability and efficiency.

Despite these advances, several challenges remain. Eye diagram-based SI evaluation is computationally expensive and not well-suited for rapid optimization \cite{expensive-eye-diagram}. Many optimization algorithms require accurate mathematical models of both the channel and equalizer, which are often difficult to obtain in practice. Furthermore, existing machine learning and RL-based approaches may require extensive training data or struggle to balance optimization effectiveness with computational efficiency, especially for complex equalizer structures.

This work addresses these challenges by proposing a purely data-driven framework for equalizer parameter optimization in high-speed DRAM systems. We develop an efficient SI metric based on learned latent representations, eliminating the need for computationally expensive eye diagram analysis. Equalizer parameter optimization is formulated as a model-free RL problem using the Advantage Actor-Critic (A2C) algorithm, with the latent representation of the signal as the state and the equalizer parameters as continuous actions. The reward is defined by the proximity of the equalized signal's latent representation to that of an ideal signal.

Our contributions are as follows: (i) a novel SI evaluation method that captures relevant signal characteristics in a learned latent space, providing both accuracy and computational efficiency; (ii) a model-free RL formulation for equalizer parameter optimization that adapts to system characteristics without requiring explicit mathematical models; and (iii) extensive experimental validation demonstrating superior performance and efficiency compared to traditional and machine learning-based methods for both DFE and cascaded CTLE+DFE structures.

The remainder of this paper is organized as follows. Section~\ref{sec:dataset} describes the DRAM waveform dataset. Section~\ref{sec:methodology} details our latent representation-based SI evaluation method and RL-based optimization framework. Section~\ref{sec:exp_setup} presents the experimental setup and baseline methods. Section~\ref{sec:results} analyzes the experimental results. Section~\ref{sec:discussion} discusses key insights and implications, and Section~\ref{sec:conclusion} concludes the paper with future research directions.

\section{Dataset}\label{sec:dataset}
The dataset used in this study comprises signal waveforms representing data transmission between the central processing unit and DRAM within a server environment operating at 6400 Mbps. These waveforms were generated through simulations of write operations, utilizing registered dual in-line memory commonly found in server systems. The dataset contains a total of 300,000 recorded values for both input and output data pairs for each of the 8 DRAMs, resulting in a total of 4.8 million samples. Each set of data gathered included two waveforms for each of the eight DRAMs: the initial input waveform written by the CPU and the output waveform, which represents the degraded signal received by the DRAM after traversing the server system components.

\begin{figure*}
	\centering
	\subfigure[Data collection setup.]
	{
		\includegraphics[width=.85\columnwidth]{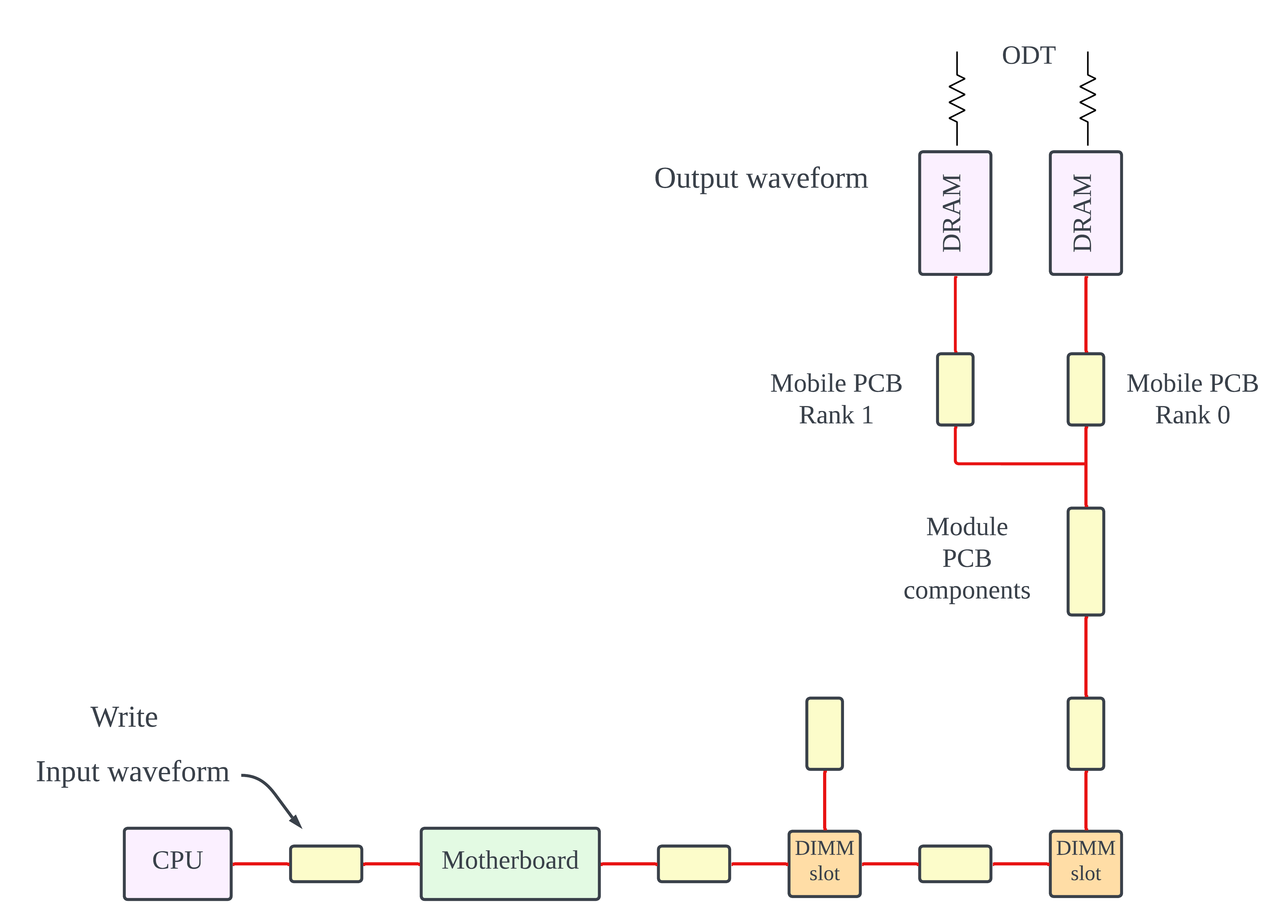}
	}
	\subfigure[Visualization of the collected dataset.]
	{
		\includegraphics[width=.85\columnwidth]{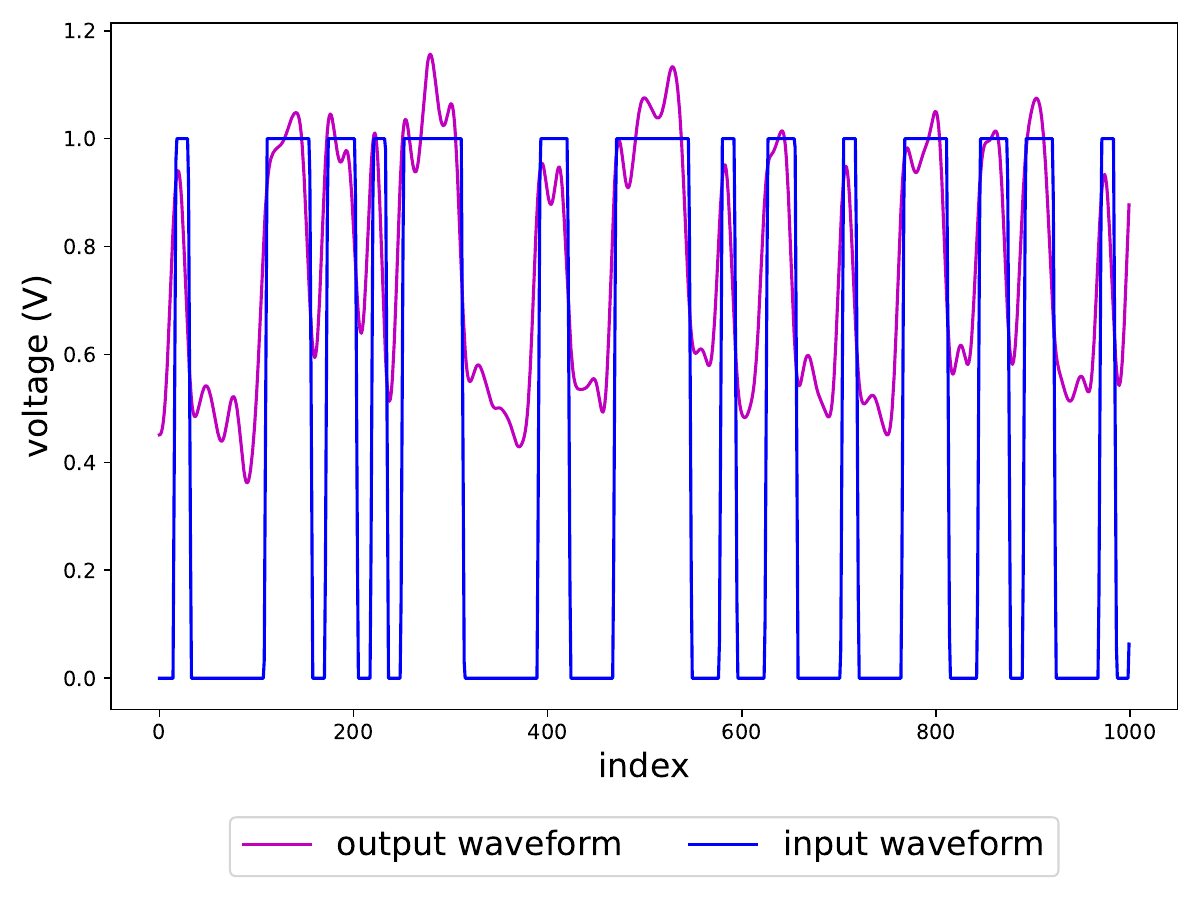}
	}
	\caption
	{
		The figure shows (a) the server memory system with double-sided DIMMs used to generate our dataset, and (b) a visualization of 1000 sample values for DRAM 1 from the dataset that plots the DRAM output waveform and the corresponding input waveform.
	}
	\label{fig:data_figures}
\end{figure*}

The dataset is organized into samples, where each sample consists of 10,000 consecutive values, \(n_x = 10000\), representing the input waveform written by the central processing unit and the corresponding output waveform received by the DRAM. Samples were constructed using a rolling window approach with a single-step increment. Both the input and output waveforms were sampled at a rate of 10 ps, with one unit interval spanning 156.3 ps. To generate eye diagrams for visual analysis, interpolation was applied to the recorded values, ensuring a time separation of 1 ps between consecutive data points. The overall data collection setup and a visualization of the dataset are shown in Fig.~\ref{fig:data_figures}.

\begin{figure*}[t]
	\centering
	\subfigure[Example of an invalid signal ($y=0$)]
	{
		\includegraphics[width=.85\columnwidth]{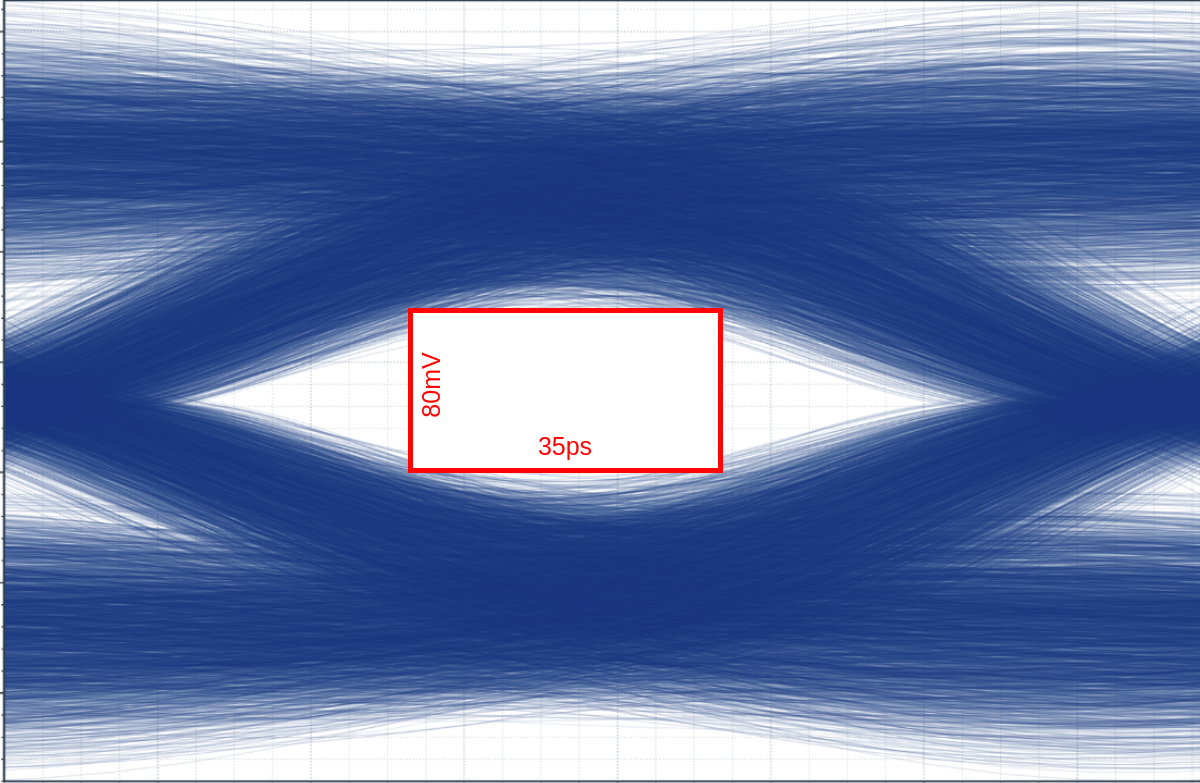}
	}
	\subfigure[Example of a valid signal ($y=1$)]
	{
		\includegraphics[width=.85\columnwidth]{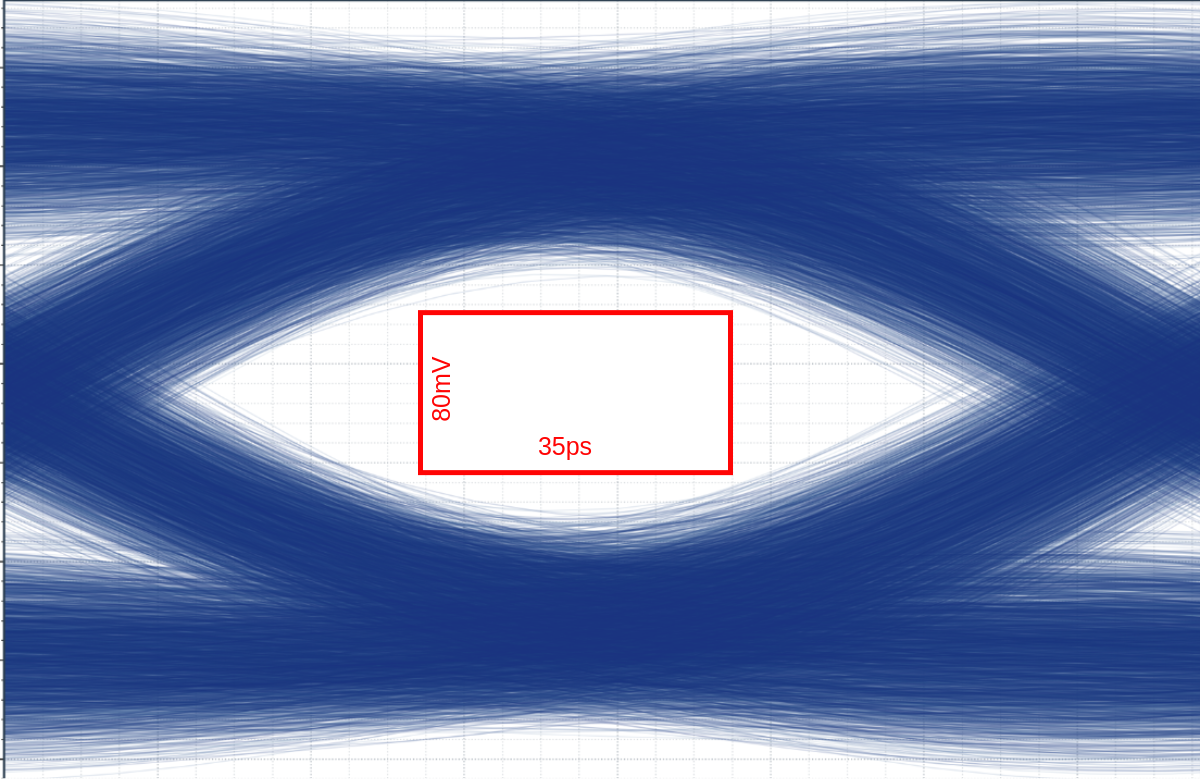}
	}
	\caption{Illustration of the signal validity labeling criteria. The rectangular window (80 mV $times$ 35 ps) is shown in red. (a) An invalid signal where signal transitions intersect the window, and (b) a valid signal where no transitions occur within the window region.}
	\label{fig:data_labeling}
\end{figure*}

Each output data sample \(d_o \in \mathbb{R}^{n_x}\) was labeled with a binary value indicating its validity, determined through eye diagram analysis. A rectangular window measuring 80 mV in height and 35 ps in width was defined within the eye-opening region. If the signal of a data sample intersects this window, it is labeled as invalid (\(y=0\)); otherwise, it is considered valid (\(y=1\)), as illustrated in Figure~\ref{fig:data_labeling}.

\section{Proposed Methodology}\label{sec:methodology}

We propose a two-stage framework for equalizer parameter optimization: (i) latent representation-based signal integrity (SI) evaluation, and (ii) reinforcement learning (RL) based equalizer optimization using the Advantage Actor-Critic (A2C) algorithm. The approach is designed for computational efficiency and robustness to channel/model uncertainties.

\subsection{Latent Representation-Based SI Evaluation}\label{sec:proposed:latent_si}

\begin{figure}[t]
	\centering
	\includegraphics[width=\linewidth]{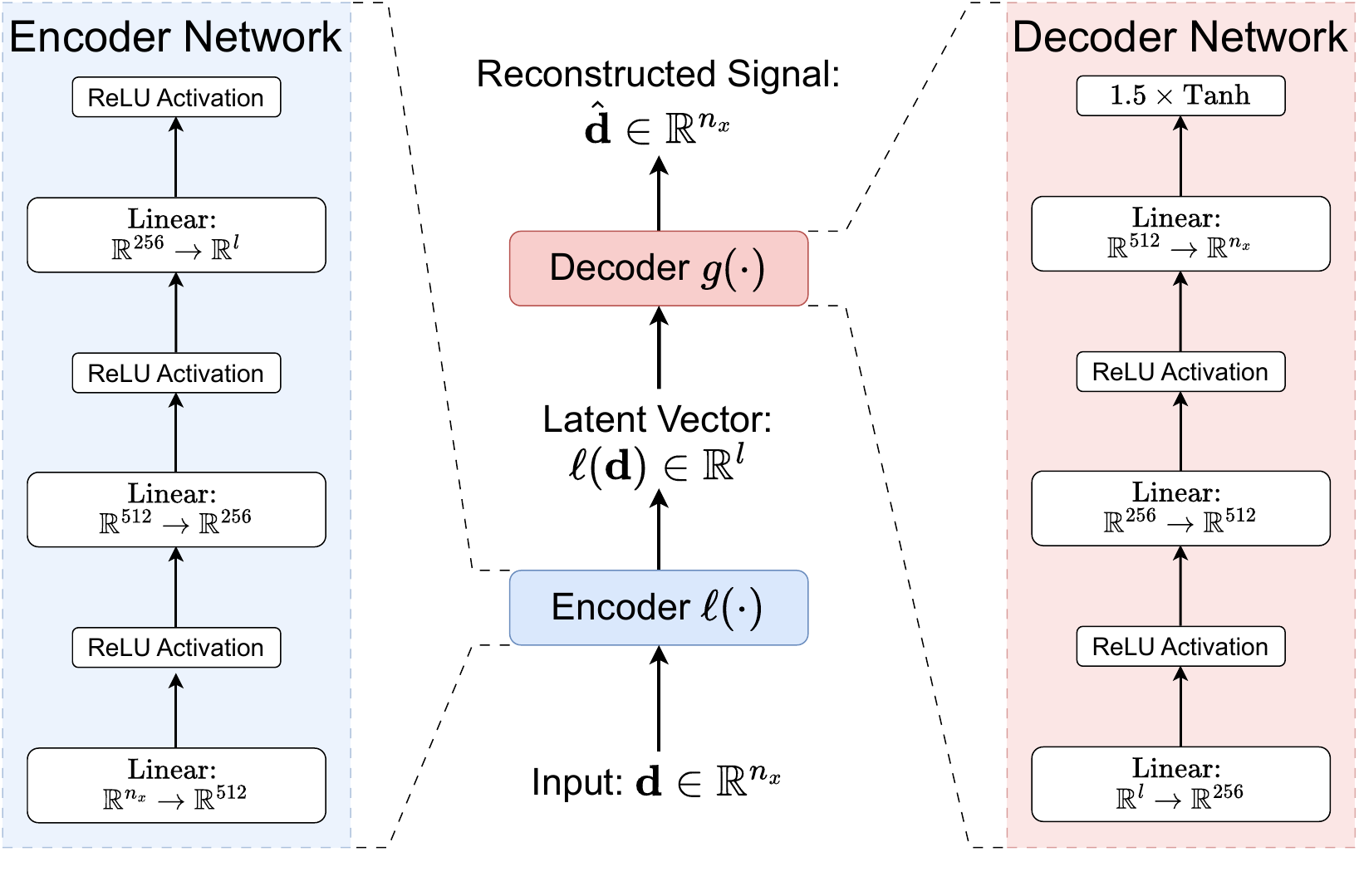}
	\caption{Autoencoder network architecture for latent SI evaluation. The encoder maps the input waveform to a latent vector via three fully connected layers with ReLU activations. The decoder reconstructs the waveform using a symmetric structure and a scaled $\tanh$ output.}
	\label{fig:autoencoder_arch}
\end{figure}

Traditional SI evaluation via eye diagrams is computationally expensive. We instead employ a learned latent space representation for rapid SI assessment. Let $\textbf{d} \in \mathbb{R}^{n_x}$ denote a waveform segment. An autoencoder with encoder $\ell(\cdot): \mathbb{R}^{n_x} \rightarrow \mathbb{R}^l$ and decoder $g(\cdot): \mathbb{R}^l \rightarrow \mathbb{R}^{n_x}$ is trained to map $\textbf{d}$ to a latent vector $\textbf{z} = \ell(\textbf{d})$. The architecture uses fully connected layers with ReLU activations and a symmetric decoder, as shown in Fig.~\ref{fig:autoencoder_arch}.

\begin{figure}[t]
    \centering
    \includegraphics[width=\linewidth]{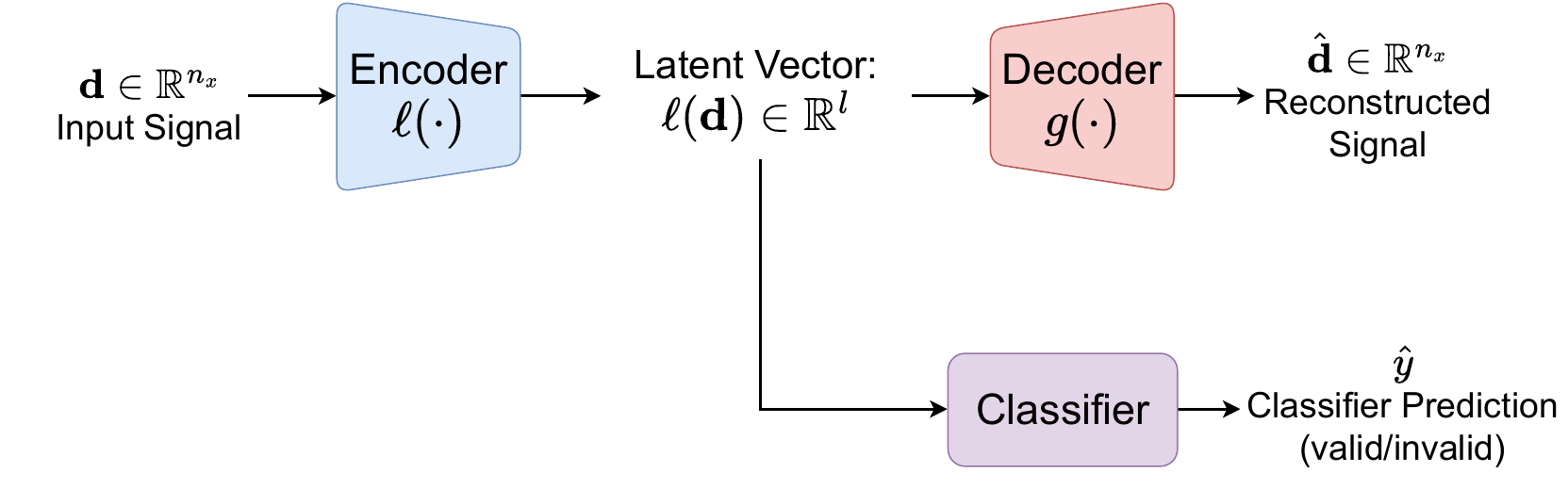}
    \caption{Autoencoder training architecture showing the encoder-decoder structure augmented with a classification head. The classifier gradients are only backpropagated for valid signals ($y=1$) to enforce tight clustering in the latent space.}
    \label{fig:autoencoder_training}
\end{figure}

To ensure SI-relevant features are captured, the reconstruction objective is augmented with a classification loss. A classifier attached to the encoder output predicts the validity of $\textbf{d}$ based on eye mask criteria. The combined loss is
\begin{equation}
	\mathcal{L} = \Vert \textbf{x} - \hat{\textbf{x}} \Vert_2^2 - y \cdot \log(\hat{y}) ,
	\label{eq:combined_loss}
\end{equation}
where $\textbf{x}$ is the input, $\hat{\textbf{x}}$ the reconstruction, $y$ the validity label, and $\hat{y}$ the classifier output. Gradients from the classifier are only backpropagated for valid signals ($y=1$), enforcing tight clustering of valid representations in latent space. The autoencoder training procedure, including the valid-only classification gradient strategy, is summarized in Algorithm~\ref{alg:latent_si}, and the overall training architecture is illustrated in Fig.~\ref{fig:autoencoder_training}.

Let $\mathbb{S} = \left\{ \ell(\textbf{d}) \in \mathbb{R}^l \mid \textbf{d} \in \mathbb{R}^{n_x},\ y = 1 \right\}$ denote the set of latent vectors corresponding to all valid signals in the dataset. The anchor point $\textbf{c} \in \mathbb{R}^l$ is computed as the Fermat-Weber point of $\mathbb{S}$:
\begin{equation}
	\textbf{c} = \argmin_{\textbf{k} \in \mathbb{S}} \sum_{j = 1}^{m} \Vert \ell(\textbf{d}_j) - \textbf{k} \Vert_2.
	\label{eq:anchor_point}
\end{equation}

\begin{algorithm}[t]
    \caption{Autoencoder Training with Valid-Only Classification Gradients}
    \label{alg:latent_si}
    \begin{algorithmic}[1]
        \Require Training dataset $\mathcal{D} = \{(\textbf{x}, y)\}$, latent dimension $l$
        \Ensure Trained encoder $\ell(\cdot)$
        \For{each batch $(\textbf{x}, y)$ from $\mathcal{D}$}
            \State $\textbf{z} \leftarrow \ell(\textbf{x})$
            \State $\hat{\textbf{x}} \leftarrow g(\textbf{z})$
            \State $\hat{y} \leftarrow c(\textbf{z})$
            \State $\mathcal{L}_r = \Vert \textbf{x} - \hat{\textbf{x}} \Vert_2^2$
            \If{$y = 1$}
                \State $\mathcal{L}_c = -y \log(\hat{y})$
                \State Update networks using $\mathcal{L}_r + \mathcal{L}_c$
            \Else
                \State Update networks using $\mathcal{L}_r$
            \EndIf
        \EndFor
        \State \Return Trained encoder $\ell(\cdot)$
    \end{algorithmic}
\end{algorithm}

\subsection{Reinforcement Learning-Based Equalizer Parameter Optimization}\label{sec:proposed:rl}

Equalizer parameter optimization is formulated as an episodic Markov Decision Process (MDP). The state space $\mathcal{S}$ comprises latent representations $\textbf{s} = \ell(\textbf{d}_o)$ of the output waveform $\textbf{d}_o$. The action space $\mathcal{A} = [0,1]^d$ consists of $d$-dimensional vectors of normalized values, where $d$ is the number of equalizer parameters (e.g., $d=4$ for DFE, $d=8$ for CTLE+DFE). These normalized actions $\textbf{a}_t \in \mathcal{A}$ are transformed into actual equalizer parameters $\textbf{p}_t$ through a predefined mapping function $M: [0,1]^d \rightarrow \mathcal{P}_{actual}$, where $\mathcal{P}_{actual}$ represents the space of valid physical parameter ranges for the specific equalizer. Thus, $\textbf{p}_t = M(\textbf{a}_t)$.

The environment transition is deterministic: applying the mapped parameters $\textbf{p}_t = M(\textbf{a}_t)$ to the current output signal $\textbf{d}_o$ yields the equalized output signal $\textbf{d}_o^e = \text{EQ}(\textbf{d}_o, \textbf{p}_t)$. The next state is obtained by encoding this equalized signal, $\textbf{s}_{t+1} = \ell(\textbf{d}_o^e)$. The reward is the negative Euclidean distance in latent space to the anchor point $\textbf{c}$:
\begin{equation}
	R(\textbf{s}_t, \textbf{a}_t) = -\Vert \textbf{c} - \ell(\text{EQ}(\textbf{d}_o, M(\textbf{a}_t))) \Vert_2,
\end{equation}
where $\text{EQ}(\cdot, \cdot)$ denotes equalization applied to the signal $\textbf{d}_o$ using the mapped parameters $M(\textbf{a}_t)$.

We use the A2C algorithm with entropy regularization. The actor parameterizes a Gaussian policy $\pi_{\boldsymbol{\theta}}(\textbf{a}|\textbf{s})$; the critic estimates $V_{\boldsymbol{\omega}}(\textbf{s})$. The advantage is
\begin{equation}
	A(\textbf{s}_t, \textbf{a}_t) = r_t + \gamma V_{\boldsymbol{\omega}}(\textbf{s}_{t+1}) - V_{\boldsymbol{\omega}}(\textbf{s}_t).
\end{equation}
The joint loss is
\begin{equation}
	\begin{aligned}
		\mathcal{L}(\boldsymbol{\theta}, \boldsymbol{\omega}) = & -\mathbb{E}_{\textbf{s}_t, \textbf{a}_t} \left[ A(\textbf{s}_t, \textbf{a}_t) \log \pi_{\boldsymbol{\theta}}(\textbf{a}_t|\textbf{s}_t) \right]\\
        & + \frac{c_v}{2} \mathbb{E}_{\textbf{s}_t} \left[ \left( r_t + \gamma V_{\boldsymbol{\omega}}(\textbf{s}_{t+1}) - V_{\boldsymbol{\omega}}(\textbf{s}_t) \right)^2 \right] \\
        & - \beta \mathbb{E}_{\textbf{s}_t} \left[ H(\pi_{\boldsymbol{\theta}}(\cdot|\textbf{s}_t)) \right],
	\end{aligned}
	\label{eq:a2c_loss}
\end{equation}
where $c_v$ and $\beta$ are weighting coefficients.

\begin{figure}[t]
	\centering
	\includegraphics[width=\linewidth]{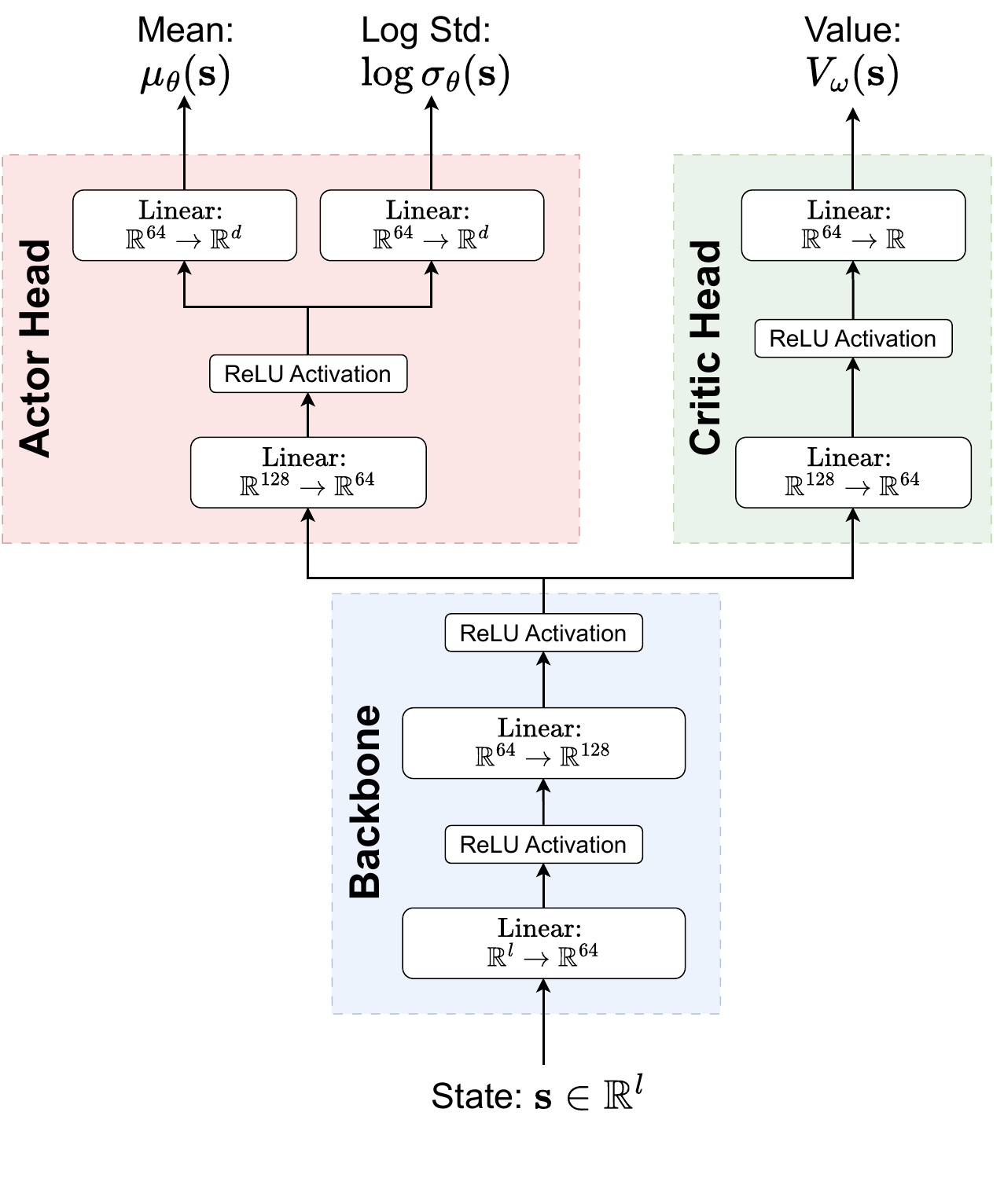}
	\caption{Network architectures for the actor and critic in the A2C framework. Both networks share a backbone of fully connected layers with ReLU activations. The actor outputs the mean and log standard deviation for the Gaussian policy; the critic outputs the state value estimate.}
	\label{fig:actor_critic_arch}
\end{figure}

The overall A2C-based optimization process is detailed in Algorithm~\ref{alg:a2c_eq}. Actor and critic networks are fully connected, taking $\textbf{s} \in \mathbb{R}^l$ as input. The actor outputs mean and log standard deviation for the Gaussian policy; the critic outputs the value estimate. The network architectures for both actor and critic are illustrated in Fig.~\ref{fig:actor_critic_arch}. During inference, the mean action is selected as the optimal parameter vector.

\begin{algorithm}[t]
	\caption{A2C-Based Equalizer Parameter Optimization}
	\label{alg:a2c_eq}
	\begin{algorithmic}[1]
		\Require Pre-trained Encoder $\ell(\cdot)$, anchor point $\textbf{c}$, training data $\mathcal{D}_{train}=\{\textbf{d}_o\}$, actor $\pi_{\boldsymbol{\theta}}$, critic $V_{\boldsymbol{\omega}}$, epochs $E$, batch size $B$, learning rate $\alpha$, discount factor $\gamma$, action mapping $M(\cdot)$
		\Ensure Trained actor $\pi_{\boldsymbol{\theta}}$
		\For{epoch $e = 1$ to $E$}
		    \State Shuffle $\mathcal{D}_{train}$ (optional)
		    \For{each batch $\{\textbf{d}_o^{(k)}\}_{k=1}^B$ from $\mathcal{D}_{train}$}
		        \For{$k = 1$ to $B$} 
		            \State $\textbf{s}_{curr}^{(k)} \leftarrow \ell(\textbf{d}_o^{(k)})$ \Comment{Get current state}
		            \State $\textbf{a}^{(k)} \sim \pi_{\boldsymbol{\theta}}(\cdot|\textbf{s}_{curr}^{(k)})$ \Comment{Sample action}
		            \State $\textbf{p}^{(k)} \leftarrow M(\textbf{a}^{(k)})$ \Comment{Map action to parameters}
		            \State $\textbf{d}_{o,eq}^{(k)} \leftarrow \text{EQ}(\textbf{d}_o^{(k)}, \textbf{p}^{(k)})$ \Comment{Equalize signal}
		            \State $\textbf{s}_{next}^{(k)} \leftarrow \ell(\textbf{d}_{o,eq}^{(k)})$ \Comment{Get next state}
		            \State $r^{(k)} \leftarrow -\Vert \textbf{c} - \textbf{s}_{next}^{(k)} \Vert_2$ \Comment{Calculate reward}
		            \State $A^{(k)} \leftarrow r^{(k)} + \gamma V_{\boldsymbol{\omega}}(\textbf{s}_{next}^{(k)}) - V_{\boldsymbol{\omega}}(\textbf{s}_{curr}^{(k)})$ \Comment{Calculate advantage}
		        \EndFor
		        \State Update $\boldsymbol{\theta}, \boldsymbol{\omega}$ using batch gradients of $\mathcal{L}(\boldsymbol{\theta}, \boldsymbol{\omega})$ from~\eqref{eq:a2c_loss}
		    \EndFor
		\EndFor
		\State \Return Trained actor $\pi_{\boldsymbol{\theta}}$
	\end{algorithmic}
\end{algorithm}

\section{Experimental Setup}\label{sec:exp_setup}
The proposed framework was evaluated on two equalizer configurations. Both equalizers were implemented in Python using NumPy. The first configuration is a 4-tap DFE. The DFE output $y[n]$ for a received signal $r[n]$ is $y[n] = r[n] - \sum_{i=1}^{4} t_i \hat{s}[n-i]$, where $t_i \in [0,1]$ are the tap weights and $\hat{s}[n-i]$ are prior hard decisions. The parameter vector is $\textbf{p}_{DFE} = \{t_1, t_2, t_3, t_4\}$, with tap weights clipped to $[0,1]$. The second configuration is a cascaded CTLE followed by a 4-tap DFE. The CTLE is modeled by $H(s) = G_{dc} \cdot \frac{s + \omega_z}{s + \omega_p} \cdot \frac{\omega_p}{\omega_z}$, with DC gain $G_{dc} \in [0,10]$, zero frequency $f_z \in [0,1]$~GHz, pole frequency $f_p \in [0,10]$~GHz, and peaking gain $G_p \in [0,20]$~dB. It is discretized using the bilinear transform. The DFE then processes the CTLE output. The parameter vector is $\textbf{p}_{CTLE+DFE} = \{G_{dc}, f_z, f_p, G_p, t_1, t_2, t_3, t_4\}$.

\begin{figure}[t]
    \centering
    \includegraphics[width=0.85\linewidth]{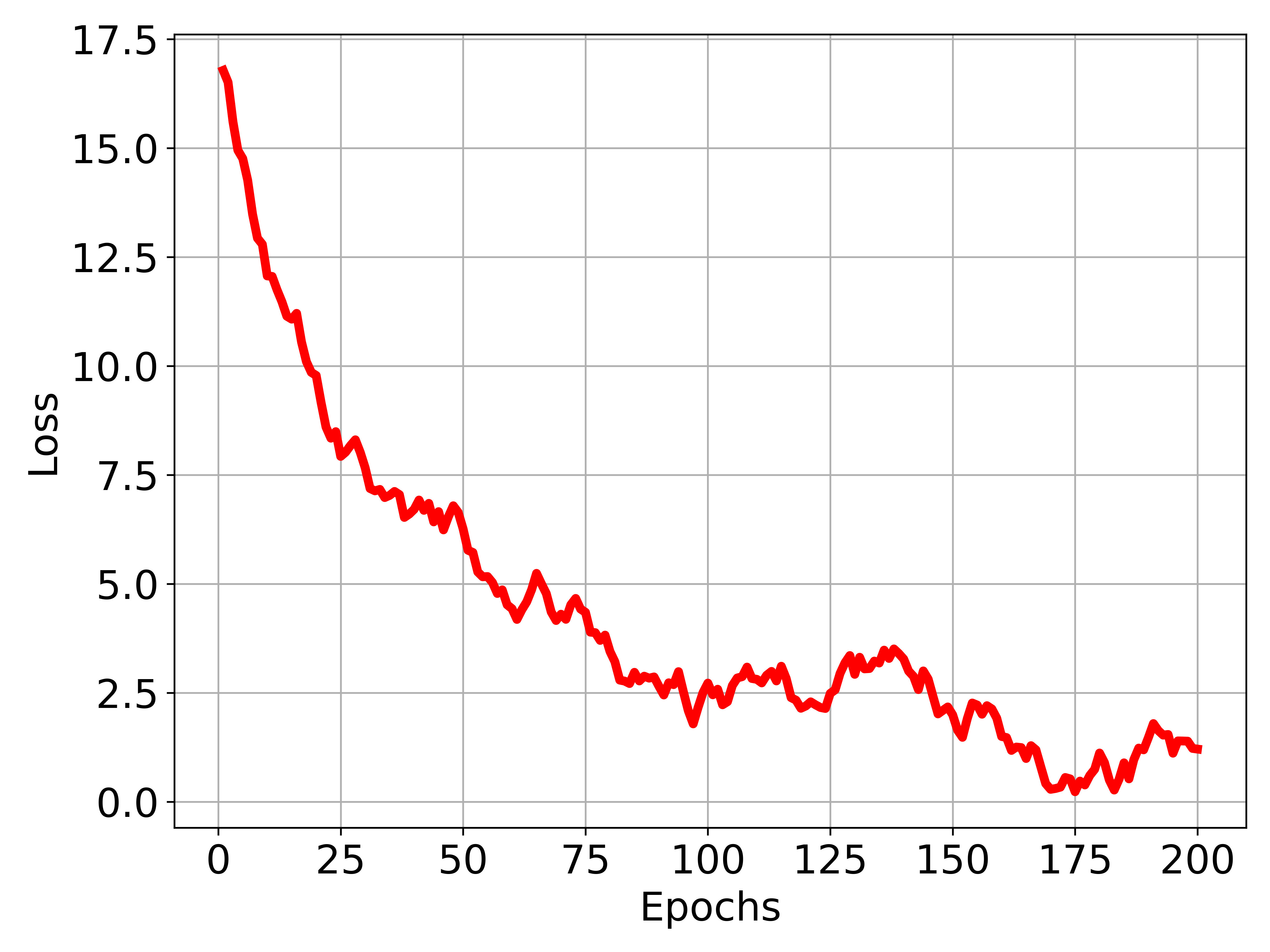}
    \caption{Training loss curve for the autoencoder network.}
    \label{fig:autoencoder_loss}
\end{figure}

The autoencoder for latent SI evaluation, detailed in Section~\ref{sec:proposed:latent_si} and Fig.~\ref{fig:autoencoder_arch}, was trained for 200 epochs using the Adam optimizer to minimize the combined loss in Eq.~\ref{eq:combined_loss}. Key hyperparameters are listed in Table~\ref{tab:hyperparameters}. The resulting training loss is shown in Fig.~\ref{fig:autoencoder_loss}.

\begin{figure}[t]
    \centering
    \includegraphics[width=0.95\linewidth]{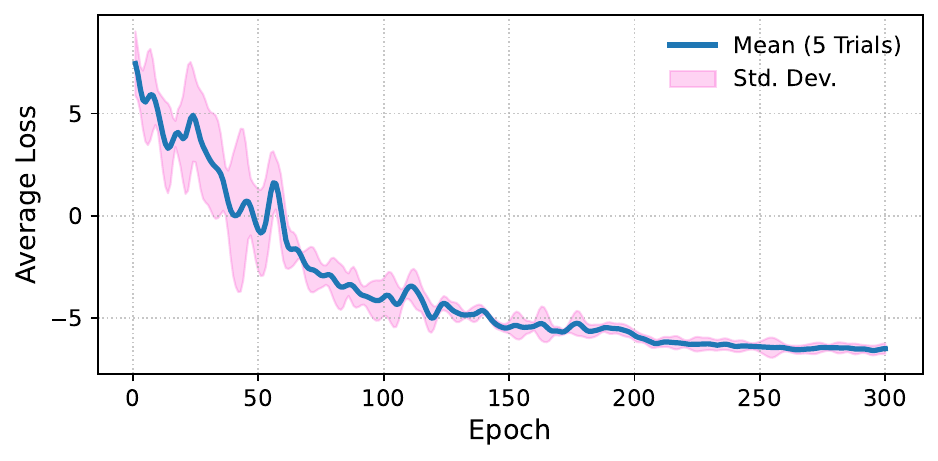}
    \caption{Training loss curve for the RL agent.}
    \label{fig:a2c_loss_curve}
\end{figure}

\begin{figure}[t]
    \centering
    \includegraphics[width=0.95\linewidth]{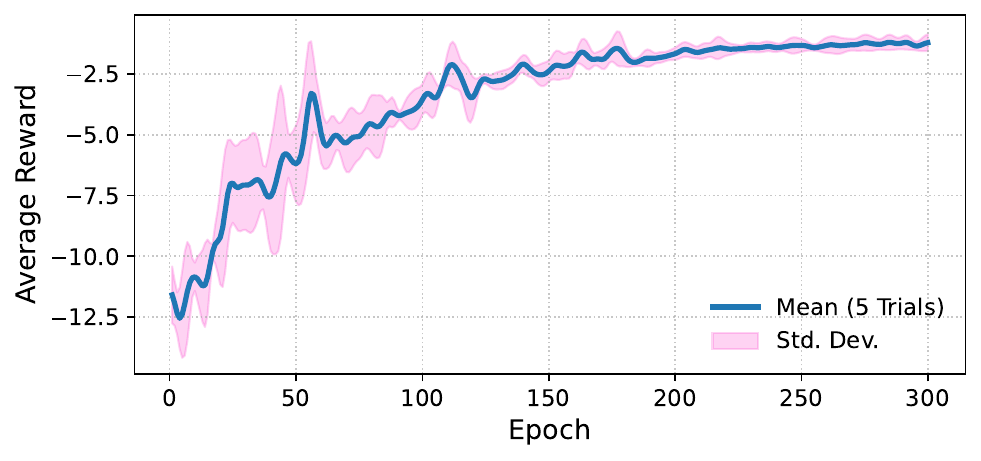}
    \caption{Training reward curve for the RL agent.}
    \label{fig:a2c_reward_curve}
\end{figure}

The A2C RL agent optimized equalizer parameters, as formulated in Section~\ref{sec:proposed:rl}. Actor and critic networks (Fig.~\ref{fig:actor_critic_arch}) utilized three fully connected layers with ReLU activations and were trained with the Adam optimizer. The RL training followed a one-step MDP formulation (Algorithm~\ref{alg:a2c_eq}) for up to 300 epochs, processing the dataset in batches, with early stopping based on reward convergence. Each batch processing constituted one training step. A2C hyperparameters are detailed in Table~\ref{tab:hyperparameters}. The training loss and reward curves for the RL agent are shown in Fig.~\ref{fig:a2c_loss_curve} and Fig.~\ref{fig:a2c_reward_curve}, respectively.

Performance was assessed by: (i) Average Window Area Improvement (\%), calculated as the percentage increase in the largest rectangular eye-opening area (via PyEye package \cite{eye-diagram-package}), and (ii) Relative Computational Time, normalized to the fastest method. Baselines included Particle Swarm Optimization (PSO), Q-learning, Bayesian optimization~\cite{ml_journal_paper}, policy optimization~\cite{baseline2}, and genetic algorithms. Further implementation details for the genetic algorithm, Bayesian optimization~\cite{ml_journal_paper}, policy optimization~\cite{baseline2}, and Q-learning are provided in Appendices~\ref{app:genetic}, \ref{app:bayesian}, \ref{app:policy}, and \ref{app:qlearning}, respectively.

\begin{table}[!t]
\renewcommand{\arraystretch}{1.3}
\label{tab:hyperparameters}
\caption{Hyperparameters for Autoencoder and RL Agent Training}
\centering
\begin{tabular}{|l|l|c|}
\hline
\textbf{Component} & \textbf{Hyperparameter} & \textbf{Value} \\
\hline
\multirow{5}{*}{Autoencoder} & Latent Dimension ($l$) & 11 \\
& Learning Rate & $1 \times 10^{-3}$ \\
& Weight Decay & $1 \times 10^{-5}$ \\
& Batch Size & 256 \\
& Epochs & 200 \\
\hline
\multirow{6}{*}{A2C RL Agent} & Learning Rate & $5 \times 10^{-4}$ \\
& Discount Factor ($\gamma$) & 0.98 \\
& Entropy Coefficient ($\beta$) & $1 \times 10^{-2}$ \\
& Value Loss Coefficient ($c_v$) & 0.5 \\
& Epochs & 300 \\
& Batch Size & 64 \\
\hline
\end{tabular}
\end{table}

\section{Results}\label{sec:results}
We evaluated our proposed methods on optimizing parameters for two equalizer structures using DRAM waveform data provided by Samsung.

\subsection{Latent Representation SI vs. Eye Diagram SI}

\begin{figure}[!t]
    \centering
    \includegraphics[width=0.48\textwidth]{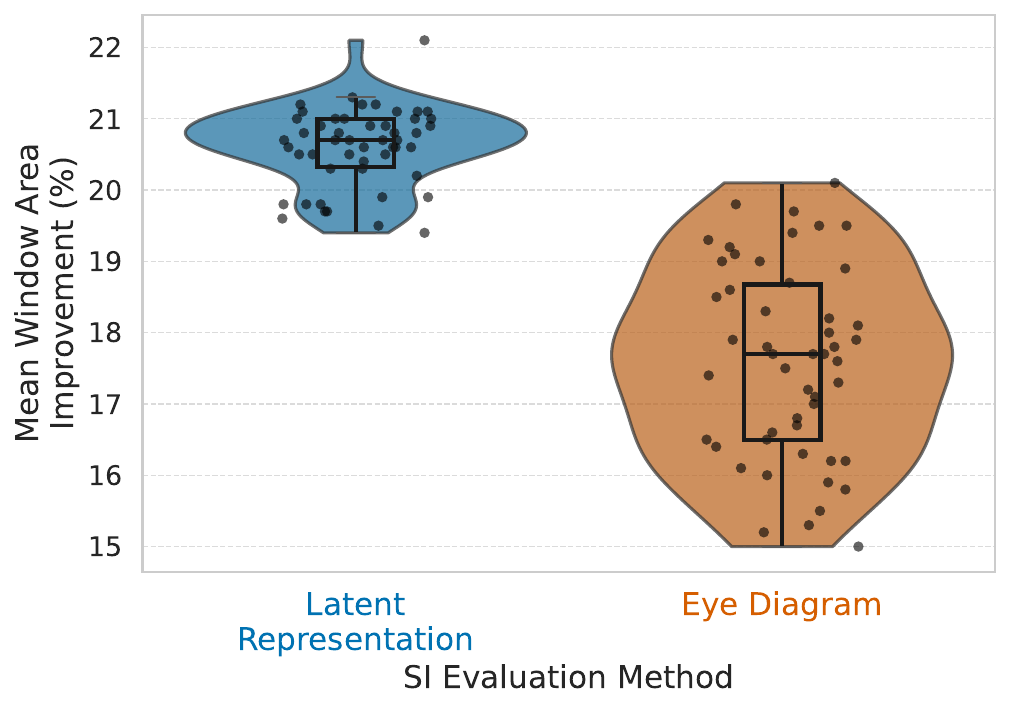}
    \caption{Distribution of mean window area improvement (\%) for DFE parameter optimization using the latent representation SI metric (blue, left) and the conventional eye diagram SI metric (orange, right) across 50 independent PSO trials. The violin plot shows the data distribution, the box plot indicates median and quartiles, and individual trial results are overlaid.}
    \label{fig:si_comparison_distribution}
\end{figure}

The latent representation-based SI metric was quantitatively compared to the conventional eye diagram-based SI evaluation for DFE parameter optimization using 50 independent PSO trials per method. As shown in Fig.~\ref{fig:si_comparison_distribution}, the latent representation approach consistently yields higher mean window area improvements with a tighter distribution. Specifically, Table~\ref{tab:si_comparison} reports an average improvement of $20.68\%$ (standard deviation $0.54\%$) for the latent method, compared to $17.54\%$ (standard deviation $1.34\%$) for the eye diagram approach. Furthermore, the latent representation metric achieves a $51\times$ reduction in computational time. These results demonstrate that the latent representation SI metric provides both superior and more consistent optimization performance with significantly greater efficiency.

\begin{table*}[!t]
\renewcommand{\arraystretch}{1.3}
\caption{Comparison of DFE Optimization Performance using Latent Representation SI vs. Eye Diagram SI (Based on 50 PSO Trials)}
\label{tab:si_comparison}
\centering
\begin{tabular}{|l|c|c|c|}
\hline

\makecell[l]{SI Evaluation\\Method} & \makecell{Average Window\\Area Improvement (\%)} & \makecell{Standard Deviation\\of Improvement (\%)} & \makecell{\makecell{Relative\\Computational Time}} \\
\hline
Latent Representation & $20.68$ & $0.54$ & $1\times$\\
Eye Diagram & $17.54$ & $1.34$ & $51\times$ \\
\hline
\end{tabular}
\end{table*}

\subsection{Results for Optimizing 4-Tap DFE Structure}

Table~\ref{tab:dfe_optimization} summarizes the quantitative results for 4-tap DFE optimization. The A2C-based RL method achieved the highest window area improvement at 36.8\% with the lowest baseline computational cost $1.0\times$. Q-learning reached 26.1\% improvement but required 7.0$\times$ more computation. PSO with latent representation yielded 19.8\% improvement at 2.0$\times$ cost. Grid search and policy optimization~\cite{baseline2} achieved 13.8\% and 15.5\% improvements, with computational costs of 8.5$\times$ and 4.0$\times$, respectively. Bayesian optimization~\cite{ml_journal_paper} and the genetic algorithm resulted in 11.7\% and 14.2\% improvements, with 5.0$\times$ and 4.5$\times$ computational time, respectively.

\begin{table}[!t]
	\renewcommand{\arraystretch}{1.3}
	\caption{Comparison of Optimization Methods for 4-Tap DFE}
	\centering
	\begin{tabular}{|l|c|c|}
		\hline
		Method                   & \makecell{Average Window                        \\Area Improvement (\%)} & \makecell{Relative\\Computational Time} \\
		\hline
		\textbf{Ours (RL - A2C)} & \textbf{36.8\%}          & \textbf{1.0$\times$} \\
		Q-learning               & 26.1                     & 7.0$\times$          \\
		PSO (Latent Rep.)        & 19.8                     & 2.0$\times$          \\
		Grid Search              & 13.8                     & 8.5$\times$          \\
		Policy Opt.~\cite{baseline2}          & 15.5                     & 4.0$\times$          \\
		Bayesian Opt.~\cite{ml_journal_paper}        & 11.7                     & 5.0$\times$          \\
		Genetic Algorithm        & 14.2                     & 4.5$\times$          \\
		\hline
	\end{tabular}
	\label{tab:dfe_optimization}
\end{table}

\subsection{Results for Optimizing Cascaded Equalizer Structure}

Table~\ref{tab:cascaded_optimization} summarizes the results for the cascaded CTLE+DFE structure. The A2C-based RL method achieved the highest window area improvement at 42.7\% with the lowest baseline computational cost $1.0\times$. Q-learning resulted in 28.5\% improvement at 13.0$\times$ computational time. PSO with latent representation achieved 25.4\% at 3.0$\times$ cost. Policy optimization~\cite{baseline2} and genetic algorithm yielded 21.3\% and 20.5\% improvements, with computational costs of 5.0$\times$ and 5.5$\times$, respectively. Grid search and Bayesian optimization~\cite{ml_journal_paper} achieved 15.2\% and 18.9\% improvements, with 32.0$\times$ and 6.5$\times$ computational time, respectively.

\begin{table}[!t]
	\renewcommand{\arraystretch}{1.3}
	\caption{Comparison of Optimization Methods for Cascaded CTLE+DFE}
	\centering
	\begin{tabular}{|l|c|c|}
		\hline
		Method                   & \makecell{Average Window                        \\Area Improvement (\%)} & \makecell{Relative\\Computational Time} \\
		\hline
		\textbf{Ours (RL - A2C)} & \textbf{42.7}            & \textbf{1.0$\times$} \\
		Q-learning               & 28.5                     & 13.0$\times$         \\
		PSO (Latent Rep.)        & 25.4                     & 3.0$\times$          \\
		Grid Search              & 15.2                     & 32.0$\times$         \\
		Policy Opt.~\cite{baseline2}          & 21.3                     & 5.0$\times$          \\
		Bayesian Opt.~\cite{ml_journal_paper}        & 18.9                     & 6.5$\times$          \\
		Genetic Algorithm        & 20.5                     & 5.5$\times$          \\
		\hline
	\end{tabular}
	\label{tab:cascaded_optimization}
\end{table}

\subsection{Visualization of Latent Space and Equalization Effect}

\begin{figure}[t]
	\centering
	\includegraphics[width=0.95\linewidth]{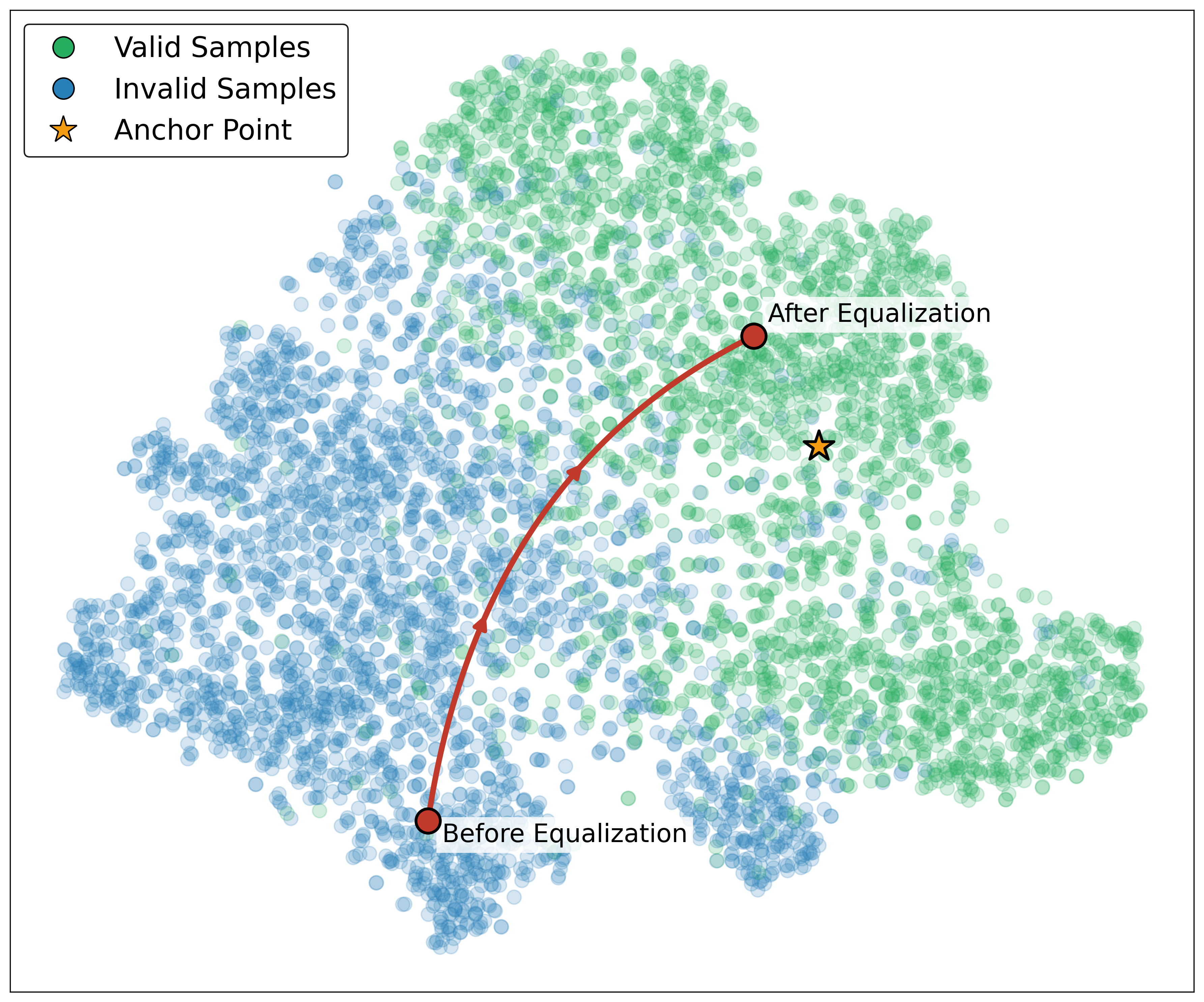}
	\caption{Two-dimensional t-SNE visualization of latent representations showing valid signals (green), invalid signals (blue), anchor point (gold star), and equalization trajectory (red line) from invalid to valid cluster.}
	\label{fig:tsne_visualization}
\end{figure}

Figure~\ref{fig:tsne_visualization} shows the two-dimensional t-SNE embeddings \cite{tsne} of signal latent representations, where valid signals form a cluster (green points), invalid signals form another cluster (blue points), and the anchor point (gold star) is positioned within the valid cluster. A sample signal's trajectory (red line) demonstrates the equalization effect: the signal starts in the invalid cluster and moves to the valid cluster after applying the optimized DFE parameters, ending near the anchor point. The clear spatial separation between valid and invalid signals in the latent space confirms the effectiveness of our SI assessment approach.

\subsection{Latent Space Dimensionality}
\begin{figure}[t]
	\centering
	\includegraphics[width=\linewidth]{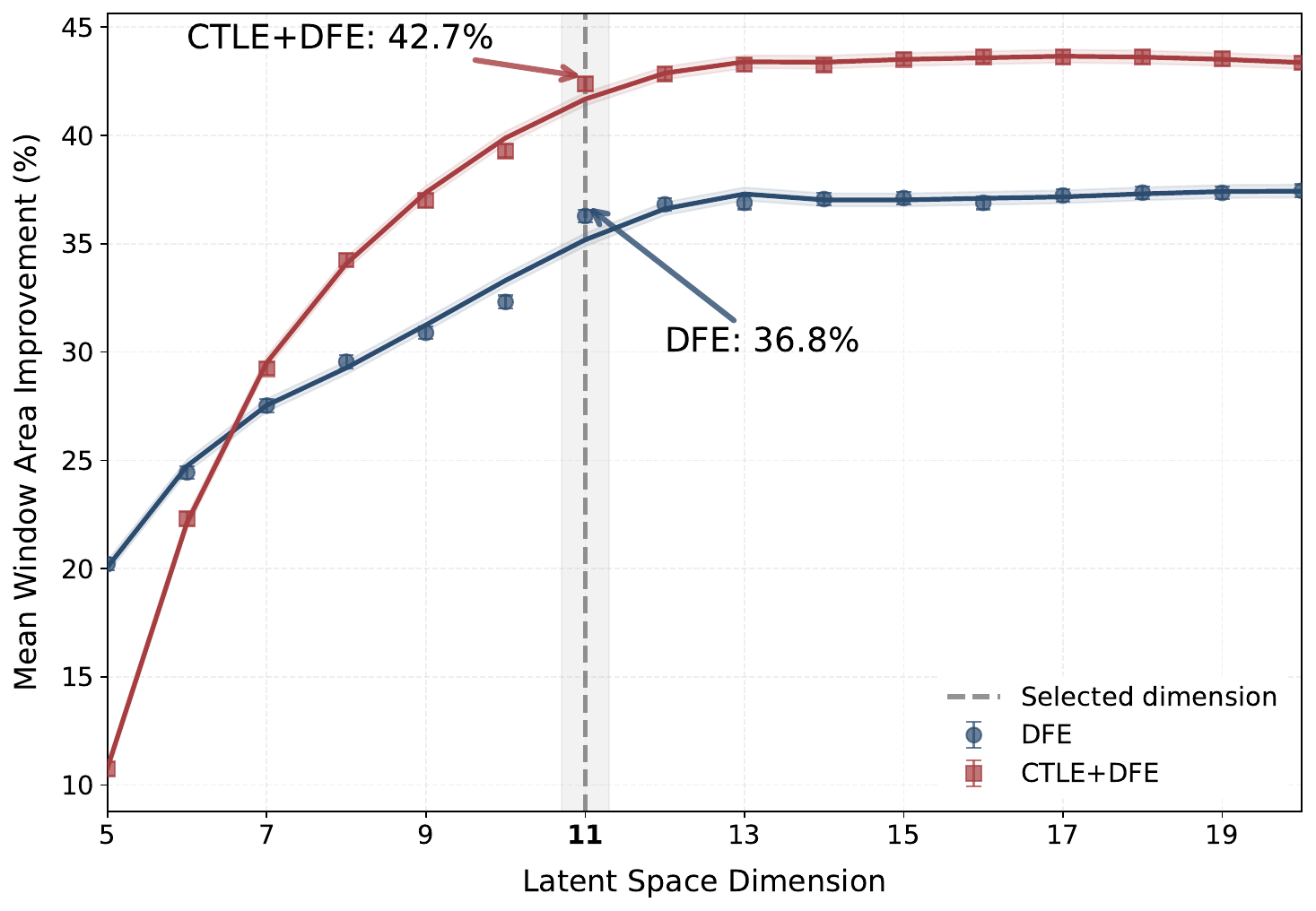}
	\caption{Mean Window area improvement versus latent space dimension. At dimension 11, DFE achieves 36.8\% and CTLE+DFE achieves 42.7\% improvement.}
	\label{fig:ablation_dim}
\end{figure}

We evaluated latent dimensions from 5 to 20 for both equalizer configurations (Figure~\ref{fig:ablation_dim}). For DFE, the mean window area improvement increased from 22.5\% (dimension 5) to 36.8\% (dimension 11), with subsequent improvements below 1.2 percentage points up to dimension 20. For CTLE+DFE, the improvement increased from its initial value to 42.7\% (dimension 11), with subsequent improvements below 0.3 percentage points. Therefore, we selected 11 as the latent dimension for our experiments.

\subsection{Generalization Across Different DRAM Units}

To evaluate the robustness of our approach and its ability to generalize, we conducted a systematic evaluation following a rigorous train-test protocol. The development phase utilized data from DRAMs 1-6, during which we trained both the autoencoder and the RL agent. These six DRAMs provided all the data needed for model training, hyperparameter tuning, and architectural decisions. DRAMs 7-8 were completely held out from this development process, serving as independent test units to assess true generalization performance.

We evaluated our trained models in two contexts: first on the familiar DRAMs 1-6 used during training, and then on the completely unseen DRAMs 7-8. Table \ref{tab:cross_dram} presents these comprehensive results. For the DFE configuration, performance on the familiar DRAMs achieved a 36.3\% improvement in window area, while the held-out DRAMs showed a 33.7\% improvement. This modest 2.6 percentage point difference suggests strong generalization. Similarly, the CTLE+DFE configuration maintained robust performance, with a 42.7\% improvement on familiar DRAMs decreasing only slightly to 39.8\% on the held-out units, representing a small 2.9 percentage point gap.

\begin{table}[!t]
    \renewcommand{\arraystretch}{1.3}
    \caption{Generalization Results: Performance Comparison Between Training and Held-Out DRAMs}
    \centering
    \begin{tabular}{|l|c|c|}
        \hline
        \multirow{2}{*}{Evaluation Set} & \multicolumn{2}{c|}{\makecell{Mean Window Area\\Improvement (\%)}} \\
        \cline{2-3}
                                        & DFE  & CTLE+DFE \\
        \hline
        Training DRAMs (1--6)           & 36.3 & 42.7     \\
        Held-Out DRAMs (7--8)           & 33.7 & 39.8     \\
        \hline
        Generalization Gap              & 2.6  & 2.9      \\
        \hline
    \end{tabular}
    \label{tab:cross_dram}
\end{table}

These results demonstrate two important characteristics of our approach. First, the CTLE+DFE configuration consistently outperforms the DFE-only structure across both familiar and unseen DRAMs, maintaining its advantage regardless of the evaluation context. Second, the small performance degradation when moving to completely new DRAM units indicates that our method has learned generally applicable optimization strategies rather than overfitting to specific unit characteristics.

The robust generalization capability shown by both equalizer configurations is particularly important for practical applications. Manufacturing variations between DRAM units are inevitable, and an optimization approach must work effectively across these variations. The minimal performance drop observed on the independent test DRAMs suggests our method can readily handle such manufacturing tolerances, making it suitable for real-world deployment.

\section{Discussion}\label{sec:discussion}
The experimental results highlight several key advantages of the proposed methodology for equalizer parameter optimization in high-speed DRAM interfaces. First, the latent representation-based signal integrity evaluation proves to be a highly efficient alternative to traditional eye diagram analysis. As detailed in Table~\ref{tab:si_comparison}, this approach not only achieves superior mean window area improvement ($20.68\%$ vs. $17.54\%$) with lower variance but also offers a significant $51\times$ reduction in computational time. The t-SNE visualization in Fig.~\ref{fig:tsne_visualization} further substantiates the efficacy of the autoencoder, illustrating a clear demarcation between valid and invalid signal clusters in the latent space and the strategic placement of the anchor point within the valid region. This confirms that the learned latent space effectively captures salient SI characteristics.

Second, the A2C reinforcement learning framework exhibits superior optimization performance compared to established baseline methods for both DFE and cascaded CTLE+DFE equalizer structures. For the 4-tap DFE, the A2C agent achieved a 36.8\% improvement in window area, surpassing Q-learning (26.1\%) and PSO with latent representation (19.8\%), while also demonstrating the lowest relative computational cost (Table~\ref{tab:dfe_optimization}). This performance advantage is even more pronounced for the more complex 8-parameter CTLE+DFE configuration, where A2C achieved a 42.7\% improvement, compared to 28.5\% for Q-learning and 25.4\% for PSO (Table~\ref{tab:cascaded_optimization}). This underscores the A2C agent's capability to effectively navigate higher-dimensional parameter spaces, a critical attribute for optimizing sophisticated equalizer designs.

Third, the proposed method demonstrates robust generalization capabilities. The performance evaluation on held-out DRAM units (DRAMs 7-8), which were not used during training, revealed only a marginal decrease in window area improvement compared to the training units (DRAMs 1-6). Specifically, the generalization gap was 2.6 percentage points for the DFE configuration and 2.9 percentage points for the CTLE+DFE configuration, as shown in Table~\ref{tab:cross_dram}. This indicates that the learned optimization policies are not overfitted to the characteristics of specific DRAM units but rather capture more generalizable features, making the approach resilient to typical manufacturing variations.

Finally, the ablation study on latent space dimensionality (Fig.~\ref{fig:ablation_dim}) identified a dimension of 11 as providing an optimal balance between model expressiveness and computational overhead. While increasing dimensionality beyond 11 yielded marginal gains in window area improvement (less than 1.2 percentage points for DFE and 0.3 percentage points for CTLE+DFE), it would incur additional computational costs. The consistent performance improvements and computational efficiency observed across different equalizer structures and DRAM units highlight the practical viability of the proposed framework for optimizing high-speed serial links.

\section{Conclusion}\label{sec:conclusion}
This work presented a data-driven framework for optimizing equalizer parameters in high-speed DRAM interfaces, combining an efficient SI evaluation metric from learned latent signal representations with a model-free A2C reinforcement learning agent. The proposed method achieved substantial eye-opening window area improvements of 42.7\% for a cascaded CTLE+DFE and 36.8\% for a DFE-only structure. These results significantly outperformed traditional and other learning-based baselines while demonstrating superior computational efficiency and robust generalization across different DRAM units.

\bibliographystyle{IEEEtran}
\bibliography{references}

\appendix

\subsection{Genetic Algorithm Implementation}\label{app:genetic}

Our genetic algorithm baseline optimizes the equalizer parameters using a population-based evolutionary approach. For the DFE configuration, it optimizes four parameters $\textbf{p} = \{t_1,t_2,t_3,t_4\}$, while for the CTLE+DFE it handles eight parameters $\textbf{p} = \{G_{dc}, f_z, f_p, G_p, t_1, t_2, t_3, t_4\}$, each within the continuous range [0, 1].

The process starts with an initial population of 25 chromosomes, randomly selected and defined by genes representing the equalizer parameters. The fitness of each chromosome is evaluated based on the window area improvement metric described in Section~\ref{sec:exp_setup}. Chromosomes with higher fitness are probabilistically selected as parents through roulette wheel selection.

For the DFE configuration, single-point crossover is applied at the second gene index, while for CTLE+DFE, two-point crossover is used at indices 3 and 6 to preserve parameter groupings. Controlled mutations, uniformly sampled in range $[-0.1, 0.1]$, are introduced to maintain diversity. Let $\zeta_{t} \in \mathbb{R}$ denote the best fitness value in the population at generation $t$. The algorithm terminates when $|\zeta_t - \zeta_{t-1}| \leq 0.0025$ for 10 consecutive generations.

\subsection{Bayesian Optimization \cite{ml_journal_paper} Baseline Implementation}\label{app:bayesian}
Our Bayesian optimization based baseline is based on the optimization method presented in \cite{ml_journal_paper}. First, we train an autoencoder network using the process described in Section \ref{sec:methodology}. Given an output signal $\textbf{d}_{o}$, we use the trained autoencoder to obtain the SI metric score, i.e. $\mu(\textbf{d}_{o})$, where $\mu(\textbf{d}_{o})$ gives the SI metric value of the signal $\textbf{d}_o$ as described in \cite{ml_journal_paper}.

For the DFE configuration, the equalizer parameters are $\textbf{p} = \{t_1,t_2,t_3,t_4\}$. For the cascaded CTLE+DFE configuration, the parameters are $\textbf{p} = \{G_{dc}, f_z, f_p, G_p, t_1, t_2, t_3, t_4\}$. These parameters $\textbf{p}$ are sampled, and the signal $\textbf{d}_{o}$ is equalized with the sampled parameters to get the equalized signal, $\textbf{d}_{o}^e$. The SI metric score for the equalized signal $\textbf{d}_{o}^e$ is calculated using the encoder network to get $\mu(\textbf{d}_{o}^e)$. The SI metric scores with and without equalization are normalized using the following procedure:
\[
	l = \frac{\mu(\textbf{d}_{o}) - \mu(\textbf{d}_{o}^e)}{\mu(\textbf{d}_{o})}.
\]
If $l > 0$, the equalization operation with parameters $\textbf{p}$ has caused an improvement in the signal integrity of the signal $\textbf{d}_o$. The objective function for the optimization problem is $\max_{\textbf{p}} l$.

For the implementation of the Bayesian optimization algorithm, we use GPflowOpt \cite{Gpflowopt}. We use Gaussian Process as our surrogate model and run the optimization process for 200 iterations. The autoencoder architecture used for this baseline is the same as our method, shown in Figure \ref{fig:autoencoder_arch}.

\subsection{Policy Optimization \cite{baseline2} Baseline Implementation}\label{app:policy}
Our policy optimization based baseline utilizes the deep deterministic policy gradient (DDPG) agent \cite{ddpg}, which learns equalizer parameters sequentially. The agent interacts with the environment (equalizer model) a number of times equal to the number of parameters to be optimized.

For the 4-tap DFE, the agent determines four parameters $\{t_1, t_2, t_3, t_4\}$ in four sequential interactions. The state $\textbf{s}_j$ at step $j \in \{1, \dots, 4\}$ consists of the first $j$ estimated parameters $\{\hat{t}_1, \dots, \hat{t}_j\}$ padded with zeros for the remaining $4-j$ parameters. The initial state is $\textbf{s}_0 = \{0,0,0,0\}$, and the terminal state is $\textbf{s}_4 = \{\hat{t}_1, \hat{t}_2, \hat{t}_3, \hat{t}_4\}$.

For the cascaded CTLE+DFE, the agent determines eight parameters $\{G_{dc}, f_z, f_p, G_p, t_1, t_2, t_3, t_4\}$ in eight sequential interactions. The state $\textbf{s}_j$ at step $j \in \{1, \dots, 8\}$ consists of the first $j$ estimated parameters padded with zeros for the remaining $8-j$ parameters. The initial state is an 8-dimensional zero vector, and the terminal state contains all eight estimated parameters.

In each interaction, the agent outputs a normalized parameter value in $[0,1]$, subsequently mapped to the actual parameter range. The actor and critic networks use the same architecture as our proposed A2C method (Fig.~\ref{fig:actor_critic_arch}). The reward $R = 100 \times (1 - \text{BER})$ is calculated after all parameters are determined. A replay memory of $50000$ and policy noise $\text{clip}(\mathcal{N}(0, 0.075^2), -0.025, 0.025)$ are used during training.

\subsection{Q-learning Baseline Implementation}\label{app:qlearning}
Our Q-learning baseline formulates equalizer parameter optimization as a one-step MDP using a Q-network with action branching architecture. The state space comprises latent representations $\ell(\textbf{d}) \in \mathbb{R}^l$ of the output signal $\textbf{d} \in \mathbb{R}^{n_x}$. The action space structure varies depending on the equalizer configuration. For the DFE case, the action space is discretized into $k=16$ levels per tap parameter, resulting in a 4-dimensional discrete space $\mathbb{Z}_k^4$, with actions mapped to tap weights through $f: \mathbb{Z}_k^4 \rightarrow [0,1]^4$. The CTLE+DFE configuration expands this to an 8-dimensional discrete space $\mathbb{Z}_k^8$ with $k=16$ levels per parameter, mapped through $f: \mathbb{Z}_k^8 \rightarrow [0,1]^8$ to obtain the full parameter set $\{G_{dc}, f_z, f_p, G_p, t_1, t_2, t_3, t_4\}$.

The Q-network architecture adapts accordingly, featuring four heads for DFE and eight heads for CTLE+DFE, where each head outputs $k$ Q-values for its corresponding parameter. Given state $\textbf{s}_t = \ell(\textbf{d}_o)$, the agent selects action $\textbf{a}_t$ using an $\epsilon$-greedy policy. The reward function $r_t = -\Vert \ell(\textbf{d}_i) - \ell(\textbf{d}_{o, \textbf{p}_t}^e) \Vert_2$ encourages equalized signals to match the ideal input signal in latent space. Network updates use temporal difference learning with target $q^{\text{target}} = r_t$, shared across all action dimensions as $\textbf{q}^{\text{target}} = r_t \cdot \mathbf{1}_m$ where $m$ equals 4 for DFE and 8 for CTLE+DFE configurations.

The training process employs a replay memory of size 50000 with batch size 128. An initial exploration rate $\epsilon = 1.0$ decays by a factor of 0.975 each epoch until reaching a minimum of 0.005. The learning rate begins at $10^{-3}$ and decreases by 10x every 25 epochs, eventually fixing at $10^{-5}$. Training concludes when the standard deviation of average rewards over a 20-epoch window falls below 0.025, indicating convergence of the learned policy.

\end{document}